\documentclass[11pt,a4paper]{article}
\usepackage[hyperref]{naaclhlt2019}
\usepackage{times}
\usepackage{latexsym}
\usepackage{booktabs}
\usepackage{amsmath}
\usepackage{amsfonts}
\usepackage{microtype}
\usepackage{color}
\usepackage{adjustbox}
\usepackage{xspace}
\usepackage{booktabs}
\usepackage{lipsum} 
\usepackage{expex}
\usepackage{mathtools}
\usepackage{caption}
\usepackage{subcaption}
\usepackage{multirow}

\usepackage{tikz}
\newcommand*\circled[1]{\tikz[baseline=(char.base)]{
            \node[shape=circle,draw,inner sep=0.85pt] (char) {#1};}}
\usepackage{xspace}
\usetikzlibrary{bayesnet}

\usepackage{url}
\usepackage{courier}  

\newcommand{\word}[1]{\textsf{\small #1}}  

\newcommand{\EOW}{\textsc{eow}\xspace}
\newcommand{\BOW}{\textsc{bow}\xspace}

\usepackage{longtable}
\usepackage[T1]{fontenc}
\usepackage[utf8]{inputenc}
\usepackage[english]{babel}
\usepackage{enumitem}

\newlist{compactdesc}{description}{2}
\setlist[compactdesc]
{topsep=0pt,partopsep=0pt,itemsep=0pt,parsep=0pt}

\usepackage[round-mode=places, round-integer-to-decimal, round-precision=1,
    table-format = 1.2,
    table-number-alignment=center,
    round-integer-to-decimal,
    output-decimal-marker={.}
    ]{siunitx}

\usepackage[disable]{todonotes}
\newcommand{\note}[4][]{\todo[author=#2,color=#3,size=\scriptsize,fancyline,caption={},#1]{#4}} 
\newcommand{\ryan}[2][]{\note[#1]{ryan}{violet!40}{#2}{}}

\newcommand{\jason}[2][]{\note[#1]{jason}{orange!40}{#2}{}}

\newcommand{\trevor}[2][]{\note[#1]{trevor}{green!40}{#2}{}}

\newcommand{\kat}[2][]{\note[#1]{kat}{green!40}{#2}{}}

\newcommand{\lex}[1]{\textit{#1}\xspace}

\newcommand{\cutforCR}[1]{}

\usepackage{cleveref}
\crefname{section}{\S}{\S\S}
\Crefname{section}{\S}{\S\S}
\crefname{table}{Tab.}{Tabs.}
\crefname{figure}{Fig.}{Figs.}
\crefname{algorithm}{Alg.}{Algs.}
\crefname{equation}{Eq.}{Eqs.}
\crefname{appendix}{App.}{Appendices}
\crefformat{section}{\S#2#1#3}  

\newcommand{\tagset}{\cal M}

\newcommand{\fs}[1]{\footnotesize{#1}}

\newcommand{\vm}{\mathbf{m}}

\newcommand{\vw}{\mathbf{w}}
\newcommand{\vc}{\mathbf{c}}

\newcommand{\vll}{\boldsymbol \ell}

\newcommand{\ww}{\mathbf{w}}

\newcommand{\cd}[1]{{\texttt{#1}}}

\newcommand{\system}[1]{\textsc{#1}\xspace}
\newcommand{\joint}{\system{joint}}
\newcommand{\gold}{\system{gold}}
\newcommand{\direct}{\system{direct}}
\newcommand{\SM}{\system{SM}}
\newcommand{\CPH}{\system{CPH}}

\newcommand\BlueF[1]{%
  \textcolor{blue}{\textit{\texttt{#1}}}
}

\newcommand\RedF[1]{%
  \textcolor{red}{\texttt{#1}}
}

\newcommand\NormalF[1]{%
  \texttt{#1}
}

\newcommand\NormalI[1]{%
  \textit{\texttt{#1}}
}

\aclfinalcopy 
\def\aclpaperid{1520} 

\title{Contextualization of Morphological Inflection}

\usepackage[safe]{tipa}
\newcommand{\spadeaff}{\textrm{\normalfont \textschwa}\xspace}
\newcommand{\clubaff}{\textrm{\normalfont \textipa{H}}\xspace}
\newcommand{\clubafff}{\textrm{\normalfont \textipa{P}}\xspace}

\author{Ekaterina Vylomova$^{\spadeaff}$~Ryan Cotterell$^{\clubafff,\clubaff}$~ Timothy Baldwin$^{\spadeaff}$~Trevor Cohn$^{\spadeaff}$~\and~Jason Eisner$^{\clubafff}$\\ \\
  $^{\spadeaff}$School of Computing and Information Systems, University of Melbourne, Melbourne, Australia \\
  $^{\clubaff}$The Computer Laboratory, University of Cambridge, Cambridge, UK\\ 
  $^{\clubafff}$Department of Computer Science, Johns Hopkins University, Baltimore, USA \\ 
  \normalsize{\{vylomovae,tbaldwin,tcohn\}@unimelb.edu.au}\\
  \normalsize{rdc42@cam.ac.uk  jason@cs.jhu.edu}}
\date{}

\begin{document}
\maketitle
\begin{abstract}
 Critical to natural language generation is the production of correctly
 inflected text. In this paper, we isolate the task of predicting a fully inflected sentence
from its partially lemmatized version.
Unlike traditional morphological inflection or surface realization, our task input does not provide ``gold'' tags that specify what morphological features to realize on each lemmatized word; rather, such features must be inferred from sentential context.
We develop a neural hybrid graphical model that explicitly reconstructs morphological features before predicting the inflected forms, and  compare this to a system
that directly predicts the inflected forms without relying on any
morphological annotation.  We experiment on several typologically
diverse languages from the Universal Dependencies treebanks, showing the utility of incorporating linguistically-motivated
latent variables into NLP models.
\end{abstract}
 \section{Introduction}
 \label{sec:introduction}
\ryan{Adjust intro to only discuss the task presented in the shared task.}
NLP systems are often required to generate grammatical text, e.g., in machine translation, summarization, dialogue, and grammar correction.  One component of grammaticality is the use of contextually appropriate 
closed-class morphemes.
In this work, we study \textbf{contextual inflection}, which has been recently introduced in the CoNLL-SIGMORPHON 2018 shared task \cite{cotterell2018conll} to directly investigate
context-dependent morphology in NLP. There, a system must inflect partially lemmatized tokens in sentential context. 
For example, in English, the system must reconstruct the correct word sequence \lex{two cats are sitting} from partially lemmatized sequence \lex{two \_cat\_ are sitting}.\trevor{The partially lemmatised example is misleading, as we fully lemmatise.} 
Among other things, this requires: (1) identifying \lex{cat} as a noun in this context, (2) recognizing that \lex{cat} should be inflected as plural to agree with the nearby verb and numeral, and (3) realizing this inflection as the suffix \lex{s}.
Most past work in supervised computational morphology---including the previous CoNLL-SIGMORPHON shared tasks on morphological reinflection \cite{cotterell-conll-sigmorphon2017}---has focused mainly on step (3) above.\looseness=-1

As the task has been introduced into the literature only recently, we provide some background. Contextual inflection amounts to a highly constrained version of language modeling.  Language modeling predicts all words of a sentence from scratch, so the usual training and evaluation metric---perplexity---is dominated by the language model's ability to predict {\em content}, which is where most of the uncertainty lies.  
Our task focuses on just the ability to reconstruct certain missing parts of the sentence---inflectional morphemes and their orthographic realization. This refocuses the modeling effort from semantic coherence to morphosyntactic coherence, an aspect of language that may take a back seat in current language models \cite[see][]{linzen2016assessing,belinkov2017neural}. 
Contextual inflection does not perfectly separate grammaticality modeling from content modeling: as illustrated in ~\cref{tab-sm}, mapping \lex{two cats \_be\_ sitting}\trevor{Again partial. Why not \lex{two cat be sit}?}  to the fully-inflected \lex{two cats were sitting} does not require full knowledge of English grammar---the system does not have to predict the required word order nor the required auxiliary verb \lex{be}, as these are supplied in the input.  Conversely, this example does still require predicting some content---the semantic choice of past tense is {\em not} given by the input and must  be guessed by the system.\looseness=-1\footnote{This morphological feature is {\em inherent} in the sense of \newcite{booij1996}.} 
\begin{table*}
\centering
  \begin{tabular}{l cccc}
    Context:  & \NormalI{two} & \NormalI{cats} & \RedF{\text{---}} & \NormalI{sitting} \\
	Lemmata:  & \NormalF{two} & \NormalF{cat} & \NormalF{be} & \NormalF{sit} \\
    Tags:  & \texttt{POS=NUM} & \begin{tabular}{@{}c@{}} \texttt{POS=NOUN} \\ \texttt{Num=Plur}\end{tabular} & \texttt{---} & \begin{tabular}{@{}c@{}c@{}} \texttt{POS=VERB} \\ \texttt{Tense=Pres} \\ \texttt{VerbForm=Part}  \end{tabular}\\
    Target:  & \BlueF{two} & \BlueF{cats} & \BlueF{were} & \BlueF{sitting} \\
  \end{tabular}
  \caption[Example data entry.]{Example data entry: the target word \lex{be} should be properly inflected into \lex{were}to fit the sentential context.}
\label{tab-sm}
\end{table*}

The primary contribution of this paper is a novel structured
neural model for contextual inflection. The model first
predicts the sequence of morphological tags from the partially lemmatized sequence and, then, it uses the predicted tag and lemma to inflect
the word. We use this model to evince
a simple point: models are better off jointly predicting morphological tags from context than directly learning to inflect lemmata from sentential context. Indeed,
none of the participants in the 2018 shared task jointly predicted tags with the inflected forms.
Comparing
our new model to several competing systems,
we show our model has the best performance on the majority of languages. We take this as
evidence that predicting morphological tags jointly with inflecting is a better method for this task. Furthermore, we provide an analysis discussing the role of morphological complexity in model performance.\looseness=-1
\section{Joint Tagging and Inflection}
Given a language, let $\tagset$ be a set of
morphological tags in accordance with the Universal Dependencies annotation \cite{nivre2016universal}. Each $m \in \tagset$ has the form $m = \langle t, \sigma \rangle$, where $t$ is a part of speech, and the slot $\sigma$ is a set of attribute--value pairs that represent morphosyntactic information, such as number, case, tense, gender, person, and others.  We take $t \in {\cal T}$, the set of universal parts of speech described by \newcite{petrov2011universal}.
%
A sentence consists of a finite word sequence $\vw$ (we use boldface for sequence variables).  For every word $w_i$ in the sequence, there is a corresponding analysis in terms of a morphological tag $m_i \in \tagset$ and a lemma $\ell_i$. 
In general, $w_i$ is determined by the
pair $\langle \ell_i, m_i \rangle$.\footnote{Although $w_i$ can sometimes be computed by concatenating $\ell_i$ with $m_i$-specific affixes, it can also be irregular.  
}
Using this notation, \newcite{cotterell2018conll}'s shared task is to predict a sentence $\vw$ from its partially lemmatized
form $\vll$, inferring $\vm$ as an intermediate latent
variable. Our dataset (\cref{sec:exp}) has all
three sequences for each sentence.%
\looseness=-1

\subsection{A Structured Neural Model}\label{sec:encoder-decoder}
\begin{figure}
  \begin{adjustbox}{width=1.\columnwidth}
    \tikzset{
  double arrow/.style args={#1 colored by #2 and #3}{
    -stealth,line width=#1,#2, 
    postaction={draw,-stealth,#3,line width=2*(#1)/3,
                shorten <=(#1)/15,shorten >=(#1)/3}, 
  }
}

\tikzstyle{arrow} = [->, >={triangle 45}]

\tikzstyle{red_arrow} = [double arrow=2pt colored by gray and red!25]
\tikzstyle{green_arrow} = [double arrow=2pt colored by gray and green!25]
\tikzstyle{blue_arrow} = [double arrow=2pt colored by gray and blue!25]

\begin{tikzpicture}
\tikzstyle{connect}=[-latex, thick]

  \node[obs] (l1) {$\ell_1$};
  \node[obs] (l2) [right=of l1] {$\ell_2$};
  \node[obs] (l3) [right=of l2] {$\ell_3$};
  \node[obs] (l4) [right=of l3] {$\ell_4$};
  
  \plate [inner sep=.10cm,yshift=.15cm] {ells} {(l1)(l2)(l3) (l4)}{$$};

  \node[latent] (m1) [below=of l1] {$m_1$};
  \node[latent] (m2) [below=of l2] {$m_2$};
  \node[latent] (m3) [below=of l3] {$m_3$};
  \node[latent] (m4) [below=of l4] {$m_4$};

  \node[latent] (f1) [below=of m1] {$w_1$};
  \node[latent] (f2) [below=of m2] {$w_2$};
  \node[latent] (f3) [below=of m3] {$w_3$};
  \node[latent] (f4) [below=of m4] {$w_4$};

  \node[factor] (fac1) [below=of l1] {};
  \node[factor] (fac2) [below=of l2] {};
  \node[factor] (fac3) [below=of l3] {};
  \node[factor] (fac4) [below=of l4] {};
  
  \node[factor] (fac5) [right=of m1] {};
  \node[factor] (fac6) [right=of m2] {};
  \node[factor] (fac7) [right=of m3] {};

  \factoredge {ells} {fac1} {};
  \factoredge {fac1} {m1} {};
  \draw [style=arrow, bend left, looseness=1.00] (l1) to (f1);
  \draw [style=arrow] (m1) to (f1);
  \factoredge {m1} {fac5} {};
  \factoredge {fac5} {m2} {};

  \factoredge {ells} {fac2} {};
  \factoredge {fac2} {m2} {};	
  \draw [style=arrow, bend left, looseness=1.00] (l2) to (f2);
  \draw [style=arrow] (m2) to (f2);
  \factoredge {m2} {fac6} {};
  \factoredge {fac6} {m3} {};

  \factoredge {ells} {fac3} {};
  \factoredge {fac3} {m3} {};	
  \draw [style=arrow, bend left, looseness=1.00] (l3) to (f3);
  \draw [style=arrow] (m3) to (f3);
  \factoredge {m3} {fac7} {};
  \factoredge {fac7} {m4} {};

  \factoredge {ells} {fac4} {};
  \factoredge {fac4} {m4} {};	
  \draw [style=arrow, bend left, looseness=1.00] (l4) to (f4);
  \draw [style=arrow] (m4) to (f4);




\end{tikzpicture}
    \end{adjustbox}
  \caption{Our structured neural model shown as a hybrid (directed--undirected) graphical model. We omitted several arcs for convenience; namely, every morphological tag $m_i$ depends on the entire sequence $\vll$. }
  \label{fig:model}
\end{figure}
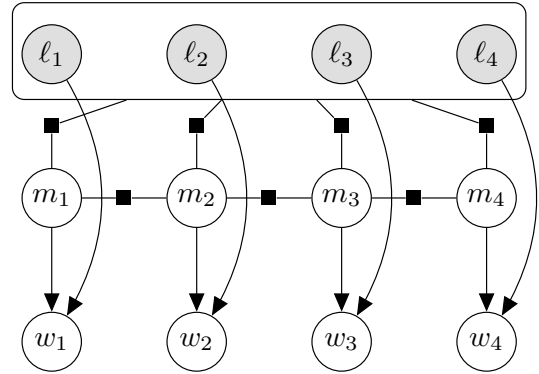

\ryan{Is it fair to call it a joint model? Discuss reviewer comments.}
Consider an extreme case when \emph{all} words are lemmatized.\footnote{In case of partially lemmatized sequence we still train the model to predict the tags over the entire sequence, but evaluate it only for lemmatized slots.}
We introduce a structured neural model for contextual
inflection, as follows:
\begin{equation}\label{eq:model}
  p(\ww, \vm \mid \vll) = \left(\prod_{i=1}^n p(w_i \mid \ell_i, m_i)
  \right) p(\vm \mid \vll)
\end{equation}
In other words, the distribution is over interleaved sequences of one-to-one
aligned inflected words and morphological tags, conditioned on a lemmatized sequence, all of length $n$. This distribution is drawn as a hybrid
(directed--undirected) graphical model \cite{Koller:2009:PGM:1795555} in \cref{fig:model}. 
We define the two
conditional distributions in the model in \cref{sec:crf} and
\cref{sec:inflector}, respectively. 

\subsection{A Neural Conditional Random Field}\label{sec:crf}
The distribution $p(\vm \mid \vll)$ is defined to be a conditional random field \cite[CRF;][]{lafferty2001conditional}.
In this work, our CRF is a conditional
distribution over morphological taggings of an input sequence. We define this conditional distribution as
\begin{equation}
p(\vm \mid \vll) = \frac{1}{Z(\vll)} \prod_{i=1}^n \psi\left(m_i, m_{i-1}, \vll\right)
\end{equation}
where $\psi(\cdot, \cdot, \cdot) \geq 0$ is an arbitrary
potential, $Z(\vll)$ normalizes the distribution, and $m_0$
is a distinguished start-of-sequence symbol. 

In this work, we opt for a recurrent neural potential---specifically,
we adopt a parameterization similar to the one given by
\newcite{lample-EtAl:2016:N16-1}. 
Our
potential $\psi$ is computed as follows. First, the sequence $\vll$ is encoded into a sequence of word vectors using the strategy described by \newcite{ling-EtAl:2015:EMNLP2}: word vectors are
passed to a bidirectional LSTM \cite{10.1007/11550907_126}, where the
corresponding hidden states are concatenated at each
time step. We simply refer to the hidden state $\mathbf{h}_i \in
\mathbb{R}^d$ as the result of said concatenation at the $i$-th
step. Using $\mathbf{h}_i$, we can define the potential function as
$  \psi\left(m_i, m_{i-1}\right) = \exp\left( A_{m_i, m_{i-1}} +
  \mathbf{o}_{m_i}^{\top} \mathbf{h}_i\right),$
where $A_{m_i, m_{i-1}}$ is a transition weight matrix and $\mathbf{o}_{m_i} \in \mathbb{R}^d$ is a morphological tag
embedding; both are learned.

\subsection{The Morphological Inflector}\label{sec:inflector}
The conditional distribution $p(w_i \mid \ell_i, m_i)$ is
parameterized by a neural encoder--decoder model with hard attention from \newcite{aharoni2016morphological}. 
The model was one of the top performers in 
the 2016 SIGMORPHON shared task \cite{cotterell-EtAl:2016:SIGMORPHON}; it achieved particularly high accuracy in the low-resource setting.
Hard attention is motivated by the observation that alignment between the input and output sequences is often monotonic in inflection tasks. 
In the model, the input lemma is treated as a sequence of characters, and encoded using a bidirectional LSTM \cite{graves2005framewise}, to produce vectors $\mathbf{x}_j$ for each character position $j$.
Next the word $w_i = \mathbf{c} = c_1\cdots c_{|w_i|}$ is generated
in a decoder character-by-character:
\begin{align}\label{eq:char-pred}
  p(c_j \mid \vc_{< j}, &l_i,m_i)= \\
  &\text{softmax}\left(\mathbf{W} \cdot \phi(\mathbf{z}_1, \ldots, \mathbf{z}_j) + \mathbf{b} \right) \nonumber
\end{align}
where $\mathbf{z}_j$ is the concatenation of the current attended input $\mathbf{x}_j$ alongside morphological features, $m_i$, and an embedding of the previously generated symbol $c_{j-1}$; and finally $\phi$ is an LSTM over the sequence of $\mathbf{z}_j$ vectors. The decoder additionally predicts a type of operation.\footnote{The model can be viewed as a transition system trained over aligned character-level strings to learn sequences of operations (\texttt{write} or \texttt{step}).}
\jason{this description overall
  is super hard to understand and needs to be rewritten, but one confusion is that you are indexing here with $i$ (actually $\mathbf{i}$), which indexes words.  Don't you mean $j$, which indexes the chars within word $i$?}  
The distribution in \cref{eq:char-pred}, strung together with the other conditionals, yields a joint distribution over the entire character sequence:\\
\begin{equation}
 p(\vc \mid \ell_i, m_i) = \prod_{j=1}^{|w_i|} p(c_j \mid  \vc_{< j}, \ell_i, m_i)
\end{equation}\ryan{Where did $\ell_i$ come from?}
\trevor{I added $\ell_i$ to conditioning on right, surely it's needed}


For instance, to map the lemma \lex{talk} to
its past form \lex{talked}, we feed in \texttt{ POS=V;Tense=PAST <w> t a l k </w>}
and train the network to output \texttt{<w> t a l k e d </w>}, where we
have augmented the orthographic character alphabet $\Sigma$ with the feature--attribute
pairs that constitute the morphological tag $m_i$.\looseness=-1 \cutforCR{ where we represent the \BOW and \EOW symbols
as \texttt{<w>} and \texttt{</w>}, respectively.}

\subsection{Parameter Estimation and Decoding}
We optimize the log-likelihood of the training data with respect to
all model parameters. As \cref{eq:model} is differentiable, this is achieved with standard gradient-based methods. 
For decoding we use a greedy
strategy where we first decode the CRF, that is, we solve the problem
$\vm^\star =
\text{argmax}_{\vm} \log p(\vm \mid \vll)$,
using the \newcite{DBLP:journals/tit/Viterbi67} algorithm. We then use
this decoded $\vm^\star$ to generate forms from the inflector.
Note that finding the one-best string under our neural inflector is intractable, and for this reason we use greedy search.


\section{Experiments}
\label{sec:exp}
\paragraph{Dataset.}
We use the Universal Dependencies v1.2 dataset
\cite{nivre2016universal} for our experiments. We include all the
languages with information on their lemmata and fine-grained grammar
tag annotation that also have \texttt{fasttext} embeddings
\cite{bojanowski2016enriching}, which are used for word embedding
initialization.\looseness=-1
\footnote{We also choose mainly non-Wikipedia datasets
  to reduce any possible intersection with the data used for the
  \textit{FastText} model training} 

\paragraph{Evaluation.}
We evaluate our model's ability to predict: (i) the correct morphological tags
from the lemma context, and (ii) the correct inflected forms. 
As our evaluation metric, we report 1-best accuracy for both
tags and word form prediction.

\paragraph{Configuration.}
We use a word and character embedding dimensionality of 300 and 100,
respectively. The hidden state dimensionality is set to 200. All models are trained with Adam \cite{kingma2014adam},
with a learning rate of 0.001 for 20 epochs. \cutforCR{No dropout was applied.}

\paragraph{Baselines.}
We use two baseline systems: (1) the CoNLL--SIGMORPHON 2018 subtask 2
neural encoder--decoder with an attention mechanism (``\SM'';
\citet{cotterell2018conll}), where the encoder represents a target form
context as a concatenation of its lemma, its left and right word forms,
their lemmata and tag representations, and then the decoder generates
the target inflected form character-by-character; and (2) a monolingual
version of the best performing system of the shared task (``\CPH'';
\citet{kementchedjhieva2018copenhagen}) that augments the above
encoder--decoder with full (sentence-level) left and right contexts (comprising of forms, their lemmata and morphological tags) as
well as predicts morphological tags for a target form as an auxiliary task.\footnote{It has been shown to improve the model's performance.} In both cases,
the hyperparameters are set as described in \citet{cotterell2018conll}.  We additionally evaluate
the SIGMORPHON baseline system on prediction of the target form without
any information on morphological tags (``\direct'').



\section{Results and Discussion}
\cref{tab-res} presents the accuracy of our best model across all
languages.\footnote{The accuracy numbers are on average higher than the ones achieved in terms of the CoNLL–SIGMORPHON 2018 subtask 2 since we did not filter out tokens that are typically not inflected (such as articles or prepositions).}
Below we highlight two main lessons from our
error analysis that apply to a wider range of generation tasks, e.g., machine translation and dialog systems.

\begin{table}
\hspace{-1ex}
\begin{adjustbox}{max width=\columnwidth}
\begin{tabular}{lcccccc}
\toprule
\multirow{2}{*}{Language} & {tag} & \multicolumn{5}{c}{form}\\
\cmidrule{3-7} 
               & \!\!\!\joint\!\!\!     & \!\!\gold\!\!         & \!\!\!\joint\!\!\!            & \!\!\!\direct\!\!\! & \!\SM\! & \!\!\CPH\!\! \\ \midrule 
\cd{Bulgarian}\!\!\!\!\!& \num{81.55}  & \num{91.89} & \num[math-rm=\mathbf]{78.81}  & \num{71.5} &\num{77.10} & \num{76.94}\\
\cd{English} & \num{89.58} &\num{95.57} & \num[math-rm=\mathbf]{90.41} & \num{86.75} & \num{86.53} & \num{86.71}\\
\cd{Basque} & \num{66.63} &\num{82.19} &\num{61.05}  & \num{59.74} & \num[math-rm=\mathbf]{61.20} & \num{60.23} \\ 
\cd{Finnish} & \num{65.99} & \num{86.53} & \num[math-rm=\mathbf]{59.34} & \num{51.21} & \num{56.61} & \num{56.40}\\
\cd{Gaelic} & \num{68.33} &\num{84.50} & \num[math-rm=\mathbf]{69.53} & \num{64.51} & \num{68.88} & \num{66.89}\\
\cd{Hindi} & \num{85.33} & \num{88.29} &\num{81.43}  & \num{85.39} & \num{86.81} & \num[math-rm=\mathbf]{87.50}\\
\cd{Italian} & \num{92.28} & \num{85.13} &\num{80.39}  & \num{85.22}  &\num{88.74} & \num[math-rm=\mathbf]{90.46}\\
\cd{Latin} & \num{82.57} & \num{89.69} &\num[math-rm=\mathbf]{75.68}  & \num{71.36} &\num{74.22} & \num{74.89}\\
\cd{Polish} & \num{71.94} & \num{96.14} & \num[math-rm=\mathbf]{74.83} & \num{61.77} & \num{72.40} & \num{70.23}\\
\cd{Swedish}& \num{81.86}  & \num{96.02} & \num[math-rm=\mathbf]{82.47} & \num{75.35} & \num{78.40} & \num{80.85}\\
\bottomrule
\end{tabular}
\end{adjustbox}
\caption{Accuracy of the models for various prediction settings. \textbf{tag}
  refers to tag prediction accuracy, and \textbf{form} 
  to form prediction accuracy. Our model is \joint; \gold denotes form prediction conditioned on gold target morphological tags; the other columns are baseline methods. }
\label{tab-res}
\end{table}
\kat{Tab2: ADD discussion: results on Italian and Hindi are interesting; both languages have high morphological tagging accuracies, but low performance on the final accuracy, even compared to the DIRECT model.}

\paragraph{Directly Predicting Morphology.}
~\cref{tab-res} indicates that all systems that make use of morphological tags outperform the \direct baseline on most languages.  
The comparison of our hybrid model with latent morphological tags to the direct form generation baseline in \SM suggests that we should be including linguistically-motivated
latent variables into models of natural language generation. We observe in \cref{tab-res} that
predicting the tag together with the form (joint) often improves performance. The most interesting comparison here is with the multi-task \CPH method, which includes morphology into the model without joint modeling; our model achieves higher results on 7/10 languages.\looseness=-1 

\paragraph{Morphological Complexity Matters.}
We observed that for languages with rich case systems, e.g.,
the Slavic languages (which exhibit a lot of fusion), the agglutinative
Finno-Ugric languages, and Basque, performance is much worse. These languages present a broader decision space and often require inferring which morphological categories need to be in agreement in order to make an accurate prediction.   
This suggests that generation in languages with more morphological
complexity will be a harder problem for neural models to solve. Indeed, this
problem is under-explored, as the field of NLP tends to fixate on generating
English text, e.g., in machine translation or dialogue system
research. \looseness=-1 



\paragraph{Error Analysis.}
We focused error analysis on prediction of agreement categories. Our analysis of adjective--noun agreement category prediction suggests that our model is able to infer adjective number, gender, and case from its head noun.   
Verb gender, which appears only
in the past tense of many Slavic languages, seems to be harder to predict. Given
that the linear distance between the subject and the verb may be longer, we
suspect the network struggles to learn longer-distance dependencies, consistent with the findings of \newcite{linzen2016assessing}. 
Overall, automatic inference of agreement categories is an interesting problem that deserves more attention, and we leave it for future work.

We also observe that most uncertainty comes from morphological
categories such as noun number, noun definiteness (which is expressed
morphologically in Bulgarian), and verb tense, all of which are inherent \cite{booij1996}\footnote{Such categories exist in most languages that exhibit some degree of morphological complexity.} and typically cannot be predicted from sentential context if they do not participate in agreement.\footnote{Unless there is no strong signal within a
sentence such as  \lex{yesterday}, \lex{tomorrow}, or \lex{ago} as in the case of tense.} On the other hand,
aspect, although being closely related to tense, is
well-predicted since it is mainly expressed as a separate
lexeme. But, in general, it is still problematic to make a prediction in languages
where aspect is morphologically marked or highly mixed with tense as in Basque.

We additionally compared 1-best and 10-best predictions for tags. Most mispredictions existing in 1-best lists are due to inherent categories mentioned above (that allow multiple plausible options that can fit the sentential context). Indeed, the problem is usually solved by allowing system to output 10-best lists. There, precision@10 is  on average 8 points higher than precision@1.

Finally, our analysis of case category prediction on nouns shows that more common cases such as the nominative, accusative, and genitive are predicted better, especially in languages with fixed word order. On the other hand, cases that appear less frequently and on shifting positions (such as the instrumental), as well as those not associated with specific prepositions, are less well predicted. In addition, we evaluated the model's performance when \emph{all} forms are replaced by their corresponding lemmata (as in \lex{two cat be sit}). For freer word order languages such as Polish or Latin, we observe a substantial drop in performance because most information on inter-word relations and their roles (expressed by means of case system) is lost.

\section{Related Work}
The primary evaluation for most contemporary language and translation
modeling research is perplexity, BLEU \cite{papineni2002bleu}, or METEOR
\cite{banerjee2005meteor}. Undoubtedly, such metrics are necessary for
extrinsic evaluation and comparison. However, relatively
few studies have focused on intrinsic evaluation of the model's
mastery of grammaticality. Recently, \newcite{linzen2016assessing}
investigated the ability of an LSTM language model to capture sentential
structure, by evaluating subject--verb agreement with respect to number, and showed
that under strong supervision, the LSTM is able to approximate dependencies.\looseness=-1 

 Taking it from the other perspective, a truer measure of grammatical competence would be a task of mapping a meaning representation to text, where the meaning representation specifies all necessary semantic content---content lemmata, dependency relations, and ``inherent'' closed-class morphemes (semantic features such as noun number, noun definiteness, and verb tense)---and the system is to realize this content according to the morphosyntactic conventions of a language, which means choosing word order, agreement morphemes, function words, and the surface forms of all words.  Such tasks have been investigated to some extent---generating text from tectogrammatical trees \cite{hajic2002natural,ptavcek2006synthesis} or from an AMR graph \cite{song-EtAl:2017:Short}. \citet{belz2011first} organized a related surface realization shared task on mapping unordered and uninflected dependency trees to properly ordered inflected sentences. The generated sentences were afterwards assessed by human annotators, making the task less scalable and more time consuming. Although our task is not perfectly matched to grammaticality modeling, the upside is that it is a ``lightweight'' task that works directly on text. No meaning representation is required. Thus, training and test data in any language can be prepared simply by lemmatizing a naturally occurring corpus.

Finally, as a morphological inflection task, the form generation
task is closely related to previous SIGMORPHON shared tasks
\cite{cotterell-EtAl:2016:SIGMORPHON,cotterell-conll-sigmorphon2017}. There,
most neural models achieve high accuracy on many languages at
type-level prediction of the form from its lemma and slot.  The
current task is more challenging in that the model has to perform token-level
form generation and inherently infer the slot from the contextual
environment. Our findings are in line with those from the CoNLL-SIGMORPHON 2018 shared task \cite{cotterell2018conll} and provide extra evidence of the utility of morphosyntactic features.




\section{Conclusion}

This work proposed a method for contextual inflection using a hybrid architecture.
Evaluation over several diverse languages showed consistent improvements over state of the art. 
Our analysis demonstrated that the contextual inflection can be a highly challenging task, and the inclusion of morphological features prediction is an important element in such a system. We also highlighted two types of morphological categories, contextual and inherent,  in which the former relies on agreement and the latter comes from a speaker's intention.

\section*{Acknowledgements}
We thank all anonymous reviewers for their comments. The first author would like to acknowledge the Google PhD fellowship. The second author would like to acknowledge a Facebook Fellowship.
\bibliography{naaclhlt2019}

\begin{thebibliography}{25}
\expandafter\ifx\csname natexlab\endcsname\relax\def\natexlab#1{#1}\fi

\bibitem[{Aharoni and Goldberg(2017)}]{aharoni2016morphological}
Roee Aharoni and Yoav Goldberg. 2017.
\newblock Morphological inflection generation with hard monotonic attention.
\newblock In \emph{Proceedings of the 55th Annual Meeting of the Association
  for Computational Linguistics (Volume 1: Long Papers)}, volume~1, pages
  2004--2015.

\bibitem[{Banerjee and Lavie(2005)}]{banerjee2005meteor}
Satanjeev Banerjee and Alon Lavie. 2005.
\newblock {METEOR}: An automatic metric for {MT} evaluation with improved
  correlation with human judgments.
\newblock In \emph{Proceedings of the {ACL} Workshop on Intrinsic and Extrinsic
  Evaluation Measures for Machine Translation and/or Summarization}, pages
  65--72.

\bibitem[{Belinkov et~al.(2017)Belinkov, Durrani, Dalvi, Sajjad, and
  Glass}]{belinkov2017neural}
Yonatan Belinkov, Nadir Durrani, Fahim Dalvi, Hassan Sajjad, and James Glass.
  2017.
\newblock What do neural machine translation models learn about morphology?
\newblock In \emph{Proceedings of the 55th Annual Meeting of the Association
  for Computational Linguistics (Volume 1: Long Papers)}, volume~1, pages
  861--872.

\bibitem[{Belz et~al.(2011)Belz, White, Espinosa, Kow, Hogan, and
  Stent}]{belz2011first}
Anja Belz, Michael White, Dominic Espinosa, Eric Kow, Deirdre Hogan, and Amanda
  Stent. 2011.
\newblock {The First Surface Realisation Shared Task}: Overview and evaluation
  results.
\newblock In \emph{Proceedings of the 13th European Workshop on Natural
  Language Generation}, pages 217--226.

\bibitem[{Bojanowski et~al.(2017)Bojanowski, Grave, Joulin, and
  Mikolov}]{bojanowski2016enriching}
Piotr Bojanowski, Edouard Grave, Armand Joulin, and Tomas Mikolov. 2017.
\newblock Enriching word vectors with subword information.
\newblock \emph{Transactions of the Association for Computational Linguistics},
  5:135--146.

\bibitem[{Booij(1996)}]{booij1996}
Geert Booij. 1996.
\newblock Inherent versus contextual inflection and the split morphology
  hypothesis authors.
\newblock In \emph{Yearbook of Morphology 1995}, pages 1--16. Springer,
  Netherlands.

\bibitem[{Cotterell et~al.(2018)Cotterell, Kirov, Sylak-Glassman, Walther,
  Vylomova, McCarthy, Kann, Mielke, Nicolai, Silfverberg, Yarowsky, Eisner, and
  Hulden}]{cotterell2018conll}
Ryan Cotterell, Christo Kirov, John Sylak-Glassman, G{\.e}raldine Walther,
  Ekaterina Vylomova, Arya~D McCarthy, Katharina Kann, Sebastian Mielke,
  Garrett Nicolai, Miikka Silfverberg, David Yarowsky, Jason Eisner, and Mans
  Hulden. 2018.
\newblock {The CoNLL--SIGMORPHON 2018 Shared Task}: Universal morphological
  reinflection.
\newblock \emph{Proceedings of the CoNLL SIGMORPHON 2018 Shared Task: Universal
  Morphological Reinflection}, pages 1--27.

\bibitem[{Cotterell et~al.(2017)Cotterell, Kirov, Sylak-Glassman, Walther,
  Vylomova, Xia, Faruqui, K{\"u}bler, Yarowsky, Eisner, and
  Hulden}]{cotterell-conll-sigmorphon2017}
Ryan Cotterell, Christo Kirov, John Sylak-Glassman, G{\'e}raldine Walther,
  Ekaterina Vylomova, Patrick Xia, Manaal Faruqui, Sandra K{\"u}bler, David
  Yarowsky, Jason Eisner, and Mans Hulden. 2017.
\newblock {The CoNLL-SIGMORPHON 2017 Shared Task}: {U}niversal morphological
  reinflection in 52 languages.
\newblock In \emph{Proceedings of the CoNLL-SIGMORPHON 2017 Shared Task:
  Universal Morphological Reinflection}.

\bibitem[{Cotterell et~al.(2016)Cotterell, Kirov, Sylak-Glassman, Yarowsky,
  Eisner, and Hulden}]{cotterell-EtAl:2016:SIGMORPHON}
Ryan Cotterell, Christo Kirov, John Sylak-Glassman, David Yarowsky, Jason
  Eisner, and Mans Hulden. 2016.
\newblock {The SIGMORPHON 2016 Shared Task}---morphological reinflection.
\newblock In \emph{Proceedings of the 14th SIGMORPHON Workshop on Computational
  Research in Phonetics, Phonology, and Morphology}, pages 10--22.

\bibitem[{Graves et~al.(2005)Graves, Fern{\'a}ndez, and
  Schmidhuber}]{10.1007/11550907_126}
Alex Graves, Santiago Fern{\'a}ndez, and J{\"u}rgen Schmidhuber. 2005.
\newblock Bidirectional {LSTM} networks for improved phoneme classification and
  recognition.
\newblock In \emph{Artificial Neural Networks: Formal Models and Their
  Applications ICANN 2005}, pages 799--804.

\bibitem[{Graves and Schmidhuber(2005)}]{graves2005framewise}
Alex Graves and J{\"u}rgen Schmidhuber. 2005.
\newblock Framewise phoneme classification with bidirectional {LSTM} and other
  neural network architectures.
\newblock \emph{Neural Networks}, 18(5-6):602--610.

\bibitem[{Hajic et~al.(2002)Hajic, Cmejrek, Dorr, Ding, Eisner, Gildea, Koo,
  Parton, Penn, Radev et~al.}]{hajic2002natural}
Jan Hajic, Martin Cmejrek, Bonnie Dorr, Yuan Ding, Jason Eisner, Daniel Gildea,
  Terry Koo, Kristen Parton, Gerald Penn, Dragomir Radev, et~al. 2002.
\newblock Natural language generation in the context of machine translation.
\newblock In \emph{Summer workshop final report, Johns Hopkins University}.

\bibitem[{Kementchedjhieva et~al.(2018)Kementchedjhieva, Bjerva, and
  Augenstein}]{kementchedjhieva2018copenhagen}
Yova Kementchedjhieva, Johannes Bjerva, and Isabelle Augenstein. 2018.
\newblock Copenhagen at {CoNLL--SIGMORPHON 2018}: Multilingual inflection in
  context with explicit morphosyntactic decoding.
\newblock \emph{CoNLL--SIGMORPHON}, page~93.

\bibitem[{Kingma and Ba(2014)}]{kingma2014adam}
Diederik~P Kingma and Jimmy Ba. 2014.
\newblock Adam: A method for stochastic optimization.
\newblock \emph{arXiv preprint arXiv:1412.6980}.

\bibitem[{Koller and Friedman(2009)}]{Koller:2009:PGM:1795555}
Daphne Koller and Nir Friedman. 2009.
\newblock \emph{Probabilistic Graphical Models: {P}rinciples and Techniques}.
\newblock The MIT Press.

\bibitem[{Lafferty et~al.(2001)Lafferty, McCallum, and
  Pereira}]{lafferty2001conditional}
John Lafferty, Andrew McCallum, and Fernando~CN Pereira. 2001.
\newblock Conditional random fields: Probabilistic models for segmenting and
  labeling sequence data.
\newblock In \emph{Proceedings of the International Conference on Machine
  Learning}.

\bibitem[{Lample et~al.(2016)Lample, Ballesteros, Subramanian, Kawakami, and
  Dyer}]{lample-EtAl:2016:N16-1}
Guillaume Lample, Miguel Ballesteros, Sandeep Subramanian, Kazuya Kawakami, and
  Chris Dyer. 2016.
\newblock Neural architectures for named entity recognition.
\newblock In \emph{Proceedings of the 2016 Conference of the North American
  Chapter of the Association for Computational Linguistics: Human Language
  Technologies}, pages 260--270.

\bibitem[{Ling et~al.(2015)Ling, Dyer, Black, Trancoso, Fermandez, Amir,
  Marujo, and Luis}]{ling-EtAl:2015:EMNLP2}
Wang Ling, Chris Dyer, Alan~W. Black, Isabel Trancoso, Ramon Fermandez, Silvio
  Amir, Luis Marujo, and Tiago Luis. 2015.
\newblock Finding function in form: {C}ompositional character models for open
  vocabulary word representation.
\newblock In \emph{Proceedings of the 2015 Conference on Empirical Methods in
  Natural Language Processing}, pages 1520--1530.

\bibitem[{Linzen et~al.(2016)Linzen, Dupoux, and
  Goldberg}]{linzen2016assessing}
Tal Linzen, Emmanuel Dupoux, and Yoav Goldberg. 2016.
\newblock Assessing the ability of {LSTMs} to learn syntax-sensitive
  dependencies.
\newblock \emph{Transactions of the Association for Computational Linguistics},
  4:521--535.

\bibitem[{Nivre et~al.(2016)Nivre, de~Marneffe, Ginter, Goldberg, Hajic,
  Manning, McDonald, Petrov, Pyysalo, Silveira et~al.}]{nivre2016universal}
Joakim Nivre, Marie-Catherine de~Marneffe, Filip Ginter, Yoav Goldberg, Jan
  Hajic, Christopher~D Manning, Ryan~T McDonald, Slav Petrov, Sampo Pyysalo,
  Natalia Silveira, et~al. 2016.
\newblock {Universal Dependencies v1}: A multilingual treebank collection.
\newblock In \emph{Proceedings of LREC 2016}.

\bibitem[{Papineni et~al.(2002)Papineni, Roukos, Ward, and
  Zhu}]{papineni2002bleu}
Kishore Papineni, Salim Roukos, Todd Ward, and Wei-Jing Zhu. 2002.
\newblock {BLEU}: A method for automatic evaluation of machine translation.
\newblock In \emph{Proceedings of the 40th Annual Meeting of the Association
  for Computational Linguistics}, pages 311--318.

\bibitem[{Petrov et~al.(2012)Petrov, Das, and McDonald}]{petrov2011universal}
Slav Petrov, Dipanjan Das, and Ryan McDonald. 2012.
\newblock A universal part-of-speech tagset.
\newblock In \emph{Proceedings of the Eighth International Conference on
  Language Resources and Evaluation (LREC-2012)}.

\bibitem[{Pt{\'a}{\v{c}}ek and
  {\v{Z}}abokrtsk{\'y}(2006)}]{ptavcek2006synthesis}
Jan Pt{\'a}{\v{c}}ek and Zden{\v{e}}k {\v{Z}}abokrtsk{\'y}. 2006.
\newblock Synthesis of {Czech} sentences from tectogrammatical trees.
\newblock In \emph{International Conference on Text, Speech and Dialogue},
  pages 221--228.

\bibitem[{Song et~al.(2017)Song, Peng, Zhang, Wang, and
  Gildea}]{song-EtAl:2017:Short}
Linfeng Song, Xiaochang Peng, Yue Zhang, Zhiguo Wang, and Daniel Gildea. 2017.
\newblock {AMR}-to-text generation with synchronous node replacement grammar.
\newblock In \emph{Proceedings of the 55th Annual Meeting of the Association
  for Computational Linguistics (Volume 2: Short Papers)}, pages 7--13.

\bibitem[{Viterbi(1967)}]{DBLP:journals/tit/Viterbi67}
Andrew~J. Viterbi. 1967.
\newblock \href {https://doi.org/10.1109/TIT.1967.1054010} {Error bounds for
  convolutional codes and an asymptotically optimum decoding algorithm}.
\newblock \emph{{IEEE} Transactions Information Theory}, 13(2):260--269.

\end{thebibliography}
\bibliographystyle{acl_natbib}

\end{document}